\documentclass[american,english]{IEEEtran}
\usepackage[latin9]{inputenc}
\usepackage{geometry}
\usepackage{babel}
\usepackage{verbatim}
\usepackage{float}
\usepackage{textcomp}
\usepackage{url}
\usepackage{amsmath}
\usepackage{amssymb}
\usepackage{graphicx}
\usepackage[unicode=true,pdfusetitle,
 bookmarks=true,bookmarksnumbered=false,bookmarksopen=false,
 breaklinks=false,pdfborder={0 0 1},backref=false,colorlinks=false]
 {hyperref}

\makeatletter

\newcommand{\noun}[1]{\textsc{#1}}
\newcommand{\lyxdot}{.}

\floatstyle{ruled}
\newfloat{algorithm}{tbp}{loa}
\providecommand{\algorithmname}{Algorithm}
\floatname{algorithm}{\protect\algorithmname}

 \let\oldforeign@language\foreign@language
 \DeclareRobustCommand{\foreign@language}[1]{%
   \lowercase{\oldforeign@language{#1}}}

\usepackage{layouts}
\usepackage{geometry}
\usepackage{graphicx}

\usepackage{dcolumn}
\usepackage{booktabs}

\usepackage{amsmath}
\usepackage{algpseudocode}
\usepackage{siunitx}
\usepackage{etoolbox}

\usepackage[table]{xcolor}
\definecolor{lightgray}{gray}{0.92}

\usepackage{multirow,bigdelim}

\usepackage{tikz}
\usetikzlibrary{arrows,shapes,matrix}

\usepackage{flushend}

\robustify\bfseries

\algrenewcommand\Return{\State \algorithmicreturn{} }%
\DeclareMathOperator*{\argmax}{arg\,max}

\algtext*{EndFor}
\algtext*{EndWhile}
\algtext*{EndIf}

\@ifundefined{showcaptionsetup}{}{%
 \PassOptionsToPackage{caption=false}{subfig}}
\usepackage{subfig}
\makeatother

\addto\captionsamerican{\renewcommand{\algorithmname}{Algorithm}}

\begin{document}
\renewcommand\figurename{Fig}\renewcommand\tablename{TABLE}\title{\ \\ \LARGE\bf Mastering $2048$ with Delayed Temporal Coherence Learning, Multi-Stage Weight Promotion, Redundant Encoding \\and Carousel Shaping \thanks{W. Ja\'skowski is with the Institute of Computing Science, Poznan University of Technology, Piotrowo 2, 60965 Pozna{\'n}, Poland and  with the Swiss AI Lab IDSIA, Galleria 2, 6928 Manno-Lugano, Switzerland email: {\tt wjaskowski@cs.put.poznan.pl}}}

\author{Wojciech Ja\'{s}kowski}

\markboth{}{Wojciech Ja\'{s}kowski: Mastering $2048$}
\maketitle
\begin{abstract}
$2048$ is an engaging single-player nondeterministic video puzzle
game, which, thanks to the simple rules and hard-to-master gameplay,
has gained massive popularity in recent years. As $2048$ can be conveniently
embedded into the discrete-state Markov decision processes framework,
we treat it as a testbed for evaluating existing and new methods in
reinforcement learning. With the aim to develop a strong $2048$ playing
program, we employ temporal difference learning with systematic $n$-tuple
networks. We show that this basic method can be significantly improved
with temporal coherence learning, multi-stage function approximator
with weight promotion, carousel shaping, and redundant encoding. In
addition, we demonstrate how to take advantage of the characteristics
of the $n$-tuple network, to improve the algorithmic effectiveness
of the learning process by delaying the (decayed) update and applying
lock-free optimistic parallelism to effortlessly make advantage of
multiple CPU cores. This way, we were able to develop the best known
$2048$ playing program to date, which confirms the effectiveness
of the introduced methods for discrete-state Markov decision problems.
\end{abstract}


\section{Introduction}

$2048$ is an engaging single-player nondeterministic puzzle game
which has gained massive popularity in recent years. Already in the
first week after its release, in March 2014, more than $500$ man-years
had been spent playing the game, according to its author. Some of
the numerous $2048$ clones\footnote{2048 itself is a derivative of the games 1024 and Threes.}
were downloaded tens of millions of times\footnote{as of March 2016}
from the online mobile app stores, making it a part of global culture.

Since the human experience with the game says that it is not trivial
to master, this raises the natural question how well a computer program
can handle it?

From the perspective of artificial intelligence, the simple rules
and hard-to-master complexity in the presence of non-determinism makes
$2048$ an ideal benchmark for learning and planning algorithms. As
$2048$ can be conveniently described in terms of Markov decision
process (MDP), in this paper, we treat $2048$ as a challenging testbed
to evaluate the effectiveness of some existing and novel ideas in
reinforcement learning for discrete-state MDPs \cite{Busoniu2010Reinforcement-L}.

For this purpose, we build upon our earlier work \cite{Szubert2014_2048},
in which we demonstrated a temporal difference (TD) learning-based
approach to $2048$. The algorithm used the standard TD($0$) rule
\cite{Sutton1998Introduction-to} to learn the weights of an $n$-tuple
network \cite{lucas2007learning} approximating the afterstate-value
function. In this paper, we extend this method in several directions.
First, we show the effectiveness of temporal coherence learning \cite{beal1999temporal},
a technique for automatically tuning the learning rates. Second, we
introduce three novel methods that concern both the learning process
and the value function approximator: i) multi-stage weight promotion,
ii) carousel shaping, and iii) redundant encoding, which are the primary
contributions of this paper. The computational experiments reveal
that the effectiveness of the first two techniques depend on the depth
of the tree the controller is allowed to search. The synergy of the
above-mentioned techniques allowed us to obtain the best $2048$ controller
to date.

This result would not be possible without paying attention to the
computational effectiveness of the learning process. In this line,
we also introduce \emph{delayed temporal difference} (delayed-TD($\lambda$)),
an online learning algorithm based on TD($\lambda$) whose performance,
when used with $n$-tuple network function approximator, is vastly
independent of the decay parameter $\lambda$. Finally, we demonstrate
that for large lookup table-based approximators such as $n$-tuple
networks, we can make use of multithreading capabilities of modern
CPUs by imposing the lock-free optimistic parallelism on the learning
process.

This work confirms the virtues of $n$-tuples network as the function
approximators. The results for $2048$ indicate that they are effective
and efficient even when involving a billion parameters, which has
never been shown before.

Although the methods introduced in this paper were developed for $2048$,
they are general, thus can be transferred to any discrete-state Markov
decision problems, classical board games included. The methodology
as a whole can be directly used to solve similar, small-board, nondeterministic
puzzle games such as $1024$, Threes, or numerous $2048$ clones.

\section{The Game of 2048}

\begin{figure}
\begin{centering}
\subfloat[A sample initial game state. All four actions are available.\label{fig:Sample-initial-game}]{\centering{}\includegraphics[width=0.38\columnwidth]{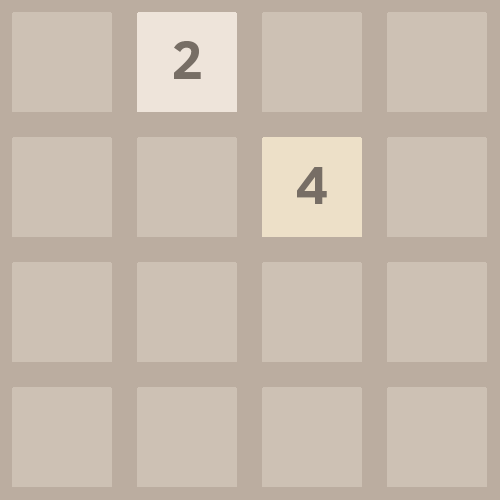}}\quad{}\subfloat[After making the \noun{up} action, a $2$-tile appeared in the upper
left corner.\label{fig:State-1}]{\centering{}\includegraphics[width=0.38\columnwidth]{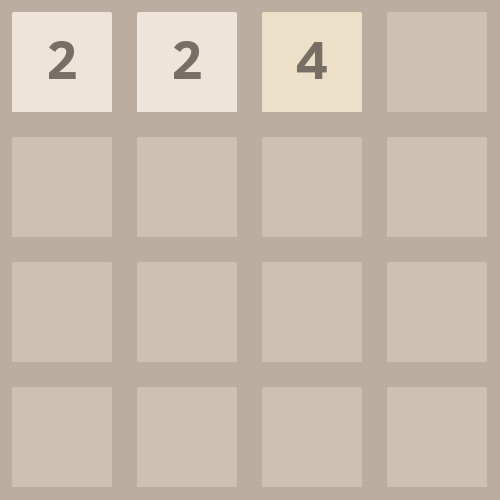}}
\par\end{centering}
\centering{}\subfloat[The \noun{left} action in $s_{1}$ was rewared by $4$ for merging
two $2$-tiles.\label{fig:State--2}]{\centering{}\includegraphics[width=0.38\columnwidth]{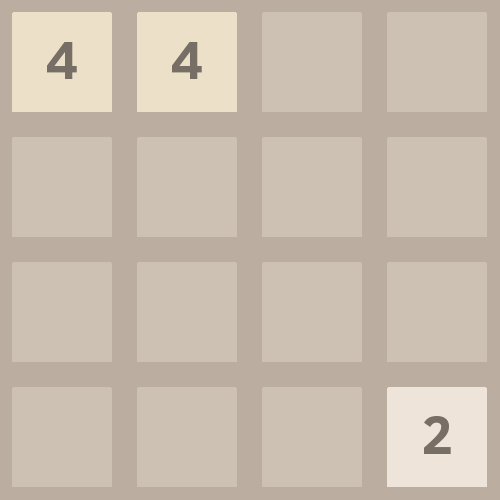}}\quad{}\subfloat[The \noun{left} action was rewarded by $8$ for merging two $4$-tiles.\label{fig:State-3}]{\centering{}\includegraphics[width=0.38\columnwidth]{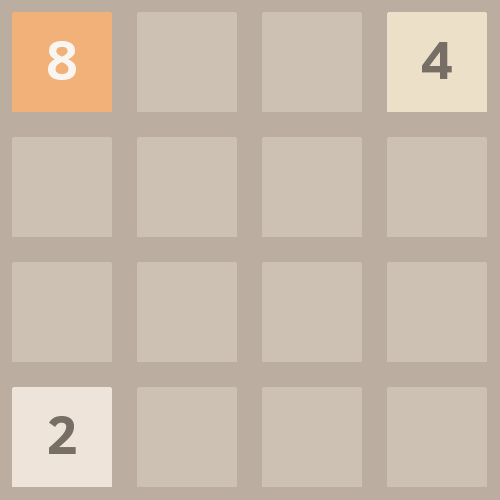}}\caption{A sample sequence of initial states and actions.}
\end{figure}
\emph{$2048$} is a single-player, nondeterministic, perfect information
video game played on a $4\times4$ board. Each square of the board
can either be empty or contain a single $v$-tile, where $v$ is a
positive power of two and denotes the value of a tile. The game starts
with two randomly generated tiles. Each time a random tile is to be
generated, a $2$-tile (with probability $p_{2}=0.9$) or a $4$-tile
($p_{4}=0.1$) is placed on an empty square of the board. A sample
initial state is shown in Fig.\,\ref{fig:Sample-initial-game}.

The objective of the game is to slide the tiles and merge the adjacent
ones to ultimately create a tile with the value of $2048$. At each
turn the player makes a move consisting in sliding all the tiles in
one of the four directions: \noun{up, right, down }or\noun{ left.
}A move is legal if at least one tile is slid. After each move, a
new $2$-tile or $4$-tile is randomly generated according to the
aforementioned probabilities. For instance, Fig.~\ref{fig:State-1}
illustrates that after sliding \noun{up} both initial tiles, a random
$2$-tile is placed in the upper left corner.

A key operation that allows obtaining tiles of larger values consists
in merging adjacent tiles. When making a move, each pair of adjacent
tiles of the same value is combined into a single tile along the move
direction. The new tile is assigned the total value of the two joined
tiles. Additionally, the player gains a reward equal to the value
of the new tile for each such merge. Figure~\ref{fig:State--2} shows
the board resulting from combining two $2$-tiles in the upper row
when sliding the tiles in the \noun{left} direction. Another \noun{left
}move leads to creating an $8$-tile (see Fig. \ref{fig:State-3}).
The moves generate rewards of $4$ and $8$, respectively.

The game is considered won when a $2048$-tile appears on the board.
However, players can continue the game even after reaching this tile.
The game terminates when no legal moves are possible, i.e., all squares
are occupied, and no two adjacent tiles are sharing the same value.
The game score is the sum of rewards obtained throughout the game.

\section{Related Work}

$2048$ was first approached by Szubert and Ja\'{s}kowski \cite{Szubert2014_2048}
who employed $\text{TD}(0)$ with $1$ million training episodes to
learn an afterstate-value function represented by a systematic $n$-tuple
network \cite{Jaskowski2014ICGAsystematic}. Their best function involved
nearly $23$ million parameters and scored $100\,178$ on average
at $1$-ply. Wu et al. \cite{wu2014multi} has later extended this
work by employing TD($0$) with $5$ million training games to learn
a larger systematic $n$-tuple system with $67$ million parameters,
scoring $142\,727$ on average. By employing three such $n$-tuple
networks enabled at different stages of the game along with expectimax
with depth $5$, they were able to achieve $328\,946$ points on average.
Further refinement its controller lead to scoring $443\,526$ on average
\cite{Yeh2016_2048}. Recently, Oka and Matsuzaki systematically evaluated
a different \emph{n}-tuple shapes to construct a $1$-ply controller
achieving $234\,136$ points on average \cite{kazuko2016_2048}. Rodgers
and Levine investigated the use of various tree search techniques
for $2048$, but were only able to score $140\,000$ on average \cite{rodgers2014investigation}.
There are also numerous $2048$ open source controllers available
on the Internet. To the best of our knowledge, the most successful
of them is by Xiao et al \cite{nneonneo}. It is based on an efficient
bitboard representation with a variable-depth expectimax, transposition
tables and a hand-designed state-evaluation heuristic, which has been
optimized by CMA-ES \cite{hansen2001completely}. Their player average
score is $442\,419$.

Mehta proved that the $n\times n$ version of the game $2048$ is
PSPACE hard \cite{mehta20142048} assuming that the random moves are
known in advance. The game has also been recently appreciated as a
tool for teaching computer science and artificial intelligence techniques
\cite{neller2015pedagogical}.

The roots of reinforcement learning \cite{kaelbling1996reinforcement}
in games can be tracked down to Samuel's famous work on Checkers \cite{samuel1959some-studies},
but it was not popularized until the Tesauro's TD-Gammon, a master-level
program for Backgammon \cite{tesauro1995temporal} obtained by temporal
difference learning. Various reinforcement learning algorithms were
applied with success to other games such as Othello \cite{Buro1995Logistello:-A-S,lucas2007learning,Szubert2013scalability},
Connect 4 \cite{Thill2012Reinforcement}, Tetris \cite{scherrer2015approximate,Jaskowski2015sztetris},
or Atari video games \cite{bellemare2012arcade}. Recently, AlphaGo
used reinforcement learning among others to determine weights for
a deep artificial neural network to beat a professional Go player
\cite{silver2016mastering}.

The $n$-tuple network is due to Bledsoe and Browning \cite{Bledsoe1959Pattern}
for optical character recognition. In the context of board games,
this idea was popularized by Lucas with its application to Othello
\cite{lucas2007learning}. It is, however, worth noticing that a similar
concept, called tabular value function, was already used by Buro in
his strong Othello-playing Logistello program \cite{Buro1995Logistello:-A-S}.
From the perspective of reinforcement learning literature, an $n$-tuple
network is a variant of tile coding \cite{Sutton1998Introduction-to},
which is another name for Albus's cerebellar model articulator controller
(CMAC) \cite{albus1971theory}.

\section{Temporal Difference Learning with N-Tuple Networks\label{sec:TDL-NTuple}}

In this section, we introduce all basic concepts for temporal difference
learning using $n$-tuple network. The learning framework presented
in the last subsection corresponds to the method we used in the previous
work \cite{Szubert2014_2048} for $2048$, and will be the foundation
for improvements in the subsequent sections.

\subsection{Markov Decision Processes}

The game of $2048$ can be conveniently framed as a fully observable,
non-discounted discrete Markov Decision Process (MDP). An MDP consists
of:
\begin{itemize}
\item a set of states $S$,
\item a set of actions available in each state $A(s)$, $s\in S$, 
\item a transition function $T(s'',s,a)$, which defines the probability
of achieving state $s''$ when making action $a$ in state $s$, and
\item a reward function $R(s)$ defining a reward received as a result of
a transition to state $s$.
\end{itemize}
An agent making decisions in MDP environment follows a policy $\pi:S\rightarrow A$,
which defines its behavior in all states. The value or utility of
a policy $\pi$ when starting the process from a given state $s\in S$
is the expected sum of rewards:
\begin{equation}
V^{\pi}(s)=\mathbb{E}_{\pi}\left[\sum_{t=0}^{\infty}R(S_{t})\vert S_{0}=s\right],\label{eq:policy-value}
\end{equation}
where $S_{t}$ is a random variable indicating the state of the process
in time $t$ and $\mathbb{E}_{\pi}$ denotes the expected value given
that the agent follows policy $\pi$. 

The goal is to find the optimal policy $\pi^{*}$ such that for all
$s\in S$
\[
\pi^{*}(s)=\mathrm{argmax}_{\pi}V^{\pi}(s).
\]

As $V^{\pi^{*}}(s)$ is the value of the optimal policy $\pi^{*}$
when following it from state $s$, it can be interpreted as a value
of state $s$, and conveniently denoted as $V(s)$. 

To solve an MDP, one can directly search the space of policies. Another
possibility is to estimate the optimal state-value function $V:S\rightarrow\mathbb{R}$.
Then, the optimal policy is a greedy policy w.r.t $V$:
\begin{equation}
\pi^{*}(s)=\mathrm{argmax}_{a\in A(s)}\sum_{s''\in S}T(s'',s,a)V(s'').\label{eq:greedy w.r.t}
\end{equation}

\subsection{Function Approximation}

For most practical problems, it is infeasible to represent a value
function as a lookup table, with one entry $V(s)$ for each state
$s\in S$, since the state space $S$ is too large. For example, the
game of 2048 has $(4\times4)^{18}\approx4.7\times10^{21}$ states,
which significantly exceeds the memory capabilities of the current
computers. This is why, in practice, instead of lookup tables, function
approximators are used. A function approximator is a function of a
form $V_{\theta}:S\rightarrow\mathbb{R}$, where $\theta$ is a vector
of parameters that undergo optimization or learning.

By using a function approximator, the dimensionality of the search
space is reduced from $|S|$ to $|\theta|$. As a result, the function
approximator gets capabilities to generalize. Thus, the choice of
a function approximator is as important as the choice of the learning
algorithm itself.

\subsubsection{$n$-Tuple Network}

One particular type of function approximator is \emph{n-tuple network
}\cite{lucas2007learning}, which has been recently successfully applied
to board games such as Othello \cite{Buro1999From-Simple-Fea,Manning2010Using-Resource-,Szubert2013On-Scalability-,Jaskowski2016cocmaes},
Connect 4 \cite{Thill2012Reinforcement}, or Tetris \cite{Jaskowski2015sztetris}.

An $n$-tuple network consists of $m$ $n_{i}$-tuples, where $n_{i}$
is tuple's size. For a given board state $s$, it calculates the sum
of values returned by the individual $n$-tuples. The $i$th $n_{i}$-tuple,
for $i=1\dots m$, consists of a predetermined sequence of board locations
$(loc_{ij})_{j=1\dots n_{i}}$, and a look-up table $V_{i}$. The
latter contains weights for each board pattern that can be observed
in the sequence of board locations. An $n$-tuple network is parametrized
by the weights in the lookup tables $(V_{i})_{i=1\dots m}$.

\begin{figure}
\centering{}\includegraphics[scale=0.8]{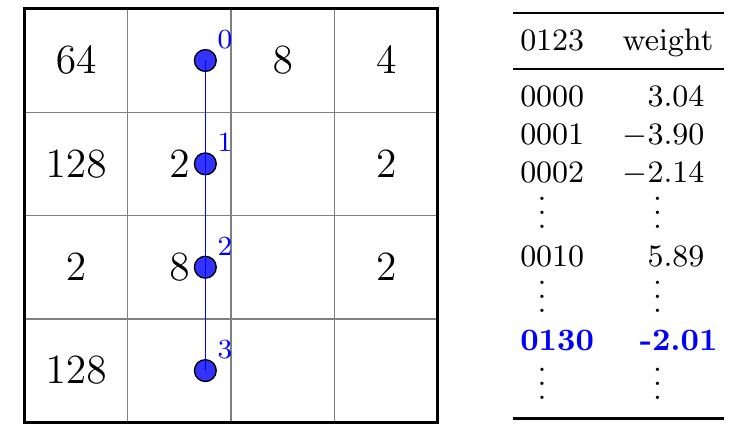} \caption{\foreignlanguage{american}{\label{fig:2048ntuples}\foreignlanguage{english}{A straight $4$-tuple
on a \emph{$2048$} board. According to the values in its lookup table,
for the given board state it returns $-2.01$, since the empty square
is encoded as $0$, the square containing $2$ as $1$, and the square
containing $8$ as $3$. }}}
\end{figure}

An $n$-tuple network is defined as follows: 
\begin{eqnarray}
V(s)=\sum_{i=1}^{m}V_{i}[index(s)] & = & \sum_{i=1}^{m}V_{i}\left[index_{i}\left(s\right)\right],\nonumber \\
index_{i}\left(s\right) & = & \sum_{j=1}^{n_{i}}s[loc_{ij}]c^{j-1},\label{eq:n-tuple-eval}
\end{eqnarray}
where $s[loc_{ij}]$ is a board value at location $loc_{ij}$, such
that $0\le s[loc_{ij}]<c$, for $j=1\dots n_{i}$, and $c$ is a constant
denoting the number of possible tile values. An empty square is encoded
as $0$, while a square containing a value $v$ as $\log_{2}v$, e.g.,
$128$ is encoded as $7$. See Fig.~\ref{fig:2048ntuples} for an
illustration.

For brevity, we will write $V_{i}[s]$ to denote $V_{i}[index_{i}(s)]$.
With this notation Eq.~\ref{eq:n-tuple-eval} boils down to 
\[
V(s)=\sum_{i=1}^{m}V_{i}[s].
\]

An $n$-tuple network is, in principle, a linear approximator, but
is based on highly nonlinear features (lookup tables). The number
of parameters of the $n$-tuple network is $O(mc^{n})$, where $m$
is the number of n-tuples, and $n$ is the size of the largest of
them. Thus, it can easily involve millions of parameters since its
number grows exponentially with $n$. At the same time, however, the
state evaluation is quick, since the algorithm defined in Eq.~\ref{eq:n-tuple-eval}
is only $O(mn)$, which make it one of the biggest advantages of $n$-tuple
networks over more complex function approximators such as artificial
neural networks. In practice, the largest factor influencing the performance
of the $n$-tuple network evaluation is the memory accesses ($s[loc_{ij}]$),
which is why we neglect (usually very small) $n$ and can claim the
time complexity to be $O(m)$.

\subsubsection{$n$-Tuple Network for $2048$}

In $2048$, there are $18$ tile values, as the tile $2^{17}$ is,
theoretically, possible to obtain. However, to limit the number of
weights in the network, we assume $c=16$. 

\begin{figure*}
\centering{}\subfloat[\foreignlanguage{american}{\label{fig:3}\foreignlanguage{english}{a straight $3$-tuple shape}}]{\includegraphics[width=0.18\textwidth]{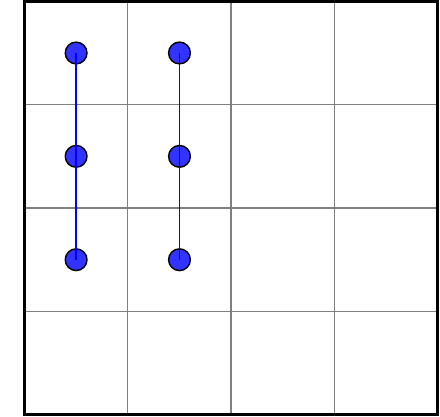}}\subfloat[\foreignlanguage{american}{\label{fig:22-4}\foreignlanguage{english}{$4$-tuple shapes: $4$
and $22$}}]{\includegraphics[width=0.18\textwidth]{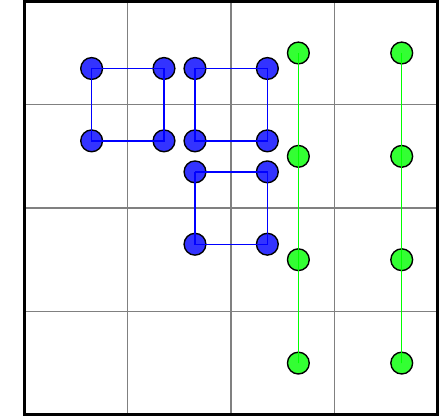}}\subfloat[\foreignlanguage{american}{\label{fig:42-33}\foreignlanguage{english}{$6$-tuple shapes: $33$
and $42$}}]{\includegraphics[width=0.18\textwidth]{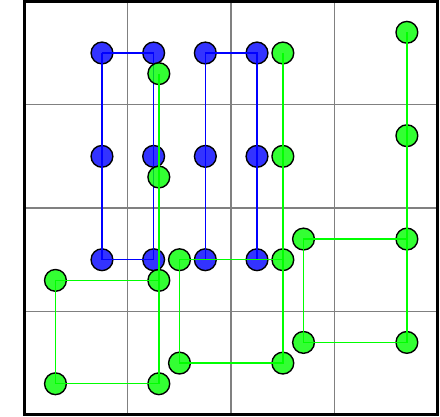}}\subfloat[\foreignlanguage{american}{\label{fig:43-421}\foreignlanguage{english}{$7$-tuple shapes: $43$
and $421$}}]{\includegraphics[width=0.18\textwidth]{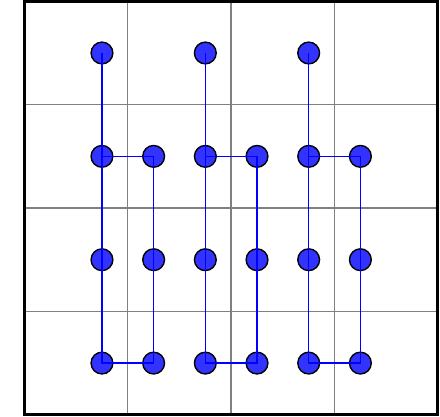}\includegraphics[width=0.18\textwidth]{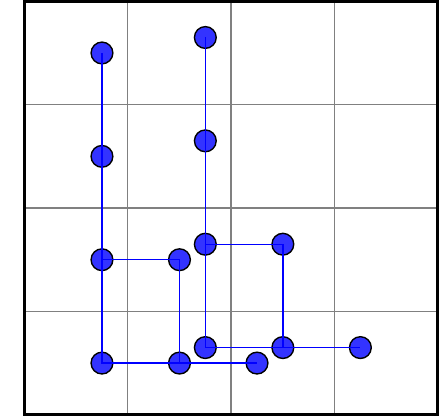}}\caption{\foreignlanguage{american}{\label{fig:NTN-architectures}\foreignlanguage{english}{$n$-tuple
shapes considered in this paper. For clarity, the symmetric expansions
were not shown.}}}
\end{figure*}
Following Wu et al. \cite{wu2014multi}, as our baseline function
approximator, we use $33$-$42$ network shown in Fig.~\ref{fig:42-33}.
Its name corresponds to the shapes of tuples used. The network consists
of four $6$-tuples, which implies $4\times16^{6}=67\,108\,864$ parameters
to learn. 

Following earlier works on $n$-tuple networks \cite{lucas2007learning,Runarsson2014Preference,Jaskowski2016cocmaes},
we also make advantage of \emph{symmetric sampling}, which is a weight
sharing technique allowing to exploit the symmetry of the board. In
symmetric sampling,\emph{ }each $n$-tuple is employed eight times
for each possible board rotation and reflection. Symmetric sampling
facilitates generalization and improves the overall performance of
the learning system without the expense of increasing the size of
the model \cite{Szubert2014_2048}.

\subsection{Temporal Difference Learning}

Temporal Difference (TD) Learning is a basic but effective state-value
function-based method to solve reinforcement learning problems. Reinforcement
learning problem is an unknown MDP, in which the learning agent does
not know the transition function or the reward function. This is not
the case in 2048, where all the game mechanics are given in advance,
but TD is perfectly applicable and has been successful also for known
MDPs.

TD updates the state-value function $V_{\theta}$ in response to the
current and next state, agent's action and its immediate consequences.
Assuming that an agent achieved state $s_{t+1}$ and reward $r_{t+1}$
by making an actions $a_{t}$ in state $s_{t}$, the TD($\lambda$)
rule uses gradient descent to update $\theta$ in the following way:
\begin{equation}
\theta_{t+1}\gets\theta_{t}+\alpha\delta_{t}\sum_{k=1}^{t}\lambda^{t-k}\nabla_{\theta_{k}}V_{\theta_{k}}(s_{k}),\label{eq:TDRuleGradient}
\end{equation}
 where the prediction error $\delta_{t}=r_{t+1}+V_{\theta_{t}}(s_{t+1})-V_{\theta_{t}}(s_{t})$,
$\alpha$ is the learning rate, and $\lambda\in[0,1]$ is the decay
parameter which determines how much of the prediction error should
be back-propagated to previously visited states. 

A special case is when $\lambda=0$. Under a convenient assumption
that $0^{0}\equiv1$, the resulting rule, called TD($0$), boils down
to:
\[
\theta_{t+1}\gets\theta_{t}+\alpha\delta_{t}\nabla_{\theta_{t}}V_{\theta_{t}}(s_{t}).
\]

\subsection{Afterstate Learning Framework with $n$-Tuple Networks}

An agent to make a decision has to evaluate all possible states it
can get into (cf. Eq.~\ref{eq:greedy w.r.t}). In $2048$, there
can be maximally $30$ such states, which can lead to a serious performance
overhead.

In the previous work \cite{Szubert2014_2048}, we have shown that
for $2048$, in order to reduce this computational burden, one can
exploit the concept of \emph{afterstates}, which are encountered in
some MDP domains. An afterstate, denoted here as $s'$, is a state
reached after the agent's action is made, but before a random element
is applied. In 2048, it is the board position obtained immediately
after sliding (and merging) the tiles. This process can be shown graphically:
\[
s'_{t}\xrightarrow{\textrm{random}}s_{t}\xrightarrow{a_{t}}s'_{t+1}\xrightarrow{\textrm{random}}s_{t+1}.
\]
The TD($\lambda$) rule for learning the afterstate-value does not
change except the fact that it involves experience tuples of $(s'_{t},a_{t},r_{t+1},s'_{t+1})$
instead of $(s_{t},a_{t},r_{t+1},s_{t+1})$, where $r_{t+1}$ is a
reward awarded for reaching state $s'_{t+1}$.

The baseline afterstate learning framework using TD($\lambda$) rule
and n-tuple network function approximator is presented in Alg.~\ref{alg:AfterstateFramework}.
The agent learns from episodes one by one. Each learning episode starts
from an initial state obtained according to the game specification
and ends when the terminal criterion is met (line $9$). The agent
follows a greedy policy w.r.t. the afterstate-value function. After
determining the action that yields the maximum afterstate value (line
$10$), the agent makes it, obtaining a reward $r_{t+1}$, an afterstate
$s'_{t+1}$ and the next state $s_{t+1}$. This information is then
used to compute the prediction error $\delta_{t}$ and to backpropagate
this error to afterstate $s'_{t}$ and its predecessors (line $13$).
Notice that in a special case when $s_{t}$ is a terminal state, the
$\delta_{t}=-V(s'_{t})$ (line $15$). 

An implementation of the TD($\lambda$) update rule for the $n$-tuple
network consisting of $m$ $n$-tuples is shown in lines $30$-$33$.
The algorithm updates the $n$-tuple network weights involved in computing
the value for state $s'_{k}$, where $k=t\dots1$. Since, the term
$\lambda^{t-k}$ decreases with decreasing $k$, which yields negligible
update values, for efficiency, we limited the update horizon to $h$
the most recently visited states. We used $h=\left\lceil \log_{\lambda}0.1\right\rceil -1$
in order to retain updates where the expression $\lambda^{t\text{\textminus}k}\ge0.1$.

It is important to observe that a standard implementation of TD($\lambda$)
involving eligibility traces \cite{Sutton1998Introduction-to} cannot
be applied to an $n$-tuple network. The eligibility traces vector
is of the same length as the weight vector. The standard Sutton's
implementation consists in updating all the elements of the vector
at each learning step. This is computationally infeasible when dealing
with hundreds of millions of weights. A potential solution would be
to maintain only the traces for activated weights \cite{thill2014temporal}.
This method has been applied to Connect Four. Notice, however, that
the number of activated weights during the episode depends on its
length. While Connect Four lasts maximally $42$ moves, $2048$ may
require several ten thousand moves. Thus, an implementation of TD($\lambda$)
requires a limited update horizon to be efficient.For convenience,
the learning rate $\alpha$ is scaled down by the number of $n$-tuples
$m$ (line $33$). As a result, $\alpha$ is independent of the number
of $n$-tuples in the network and it is easier to interpret: $\alpha=1.0$
is the maximum sensible value, since for $s'_{t}$, it immediately
reduces the prediction error $\delta_{t}$ to zero ($V_{t+1}(s'_{t})\gets r_{t+1}+V_{t}(s'_{t+1})$).

This learning framework does not include exploration, i.e., making
non-greedy actions. Despite experimenting with $\epsilon$-greedy,
softmax, and other custom exploration ideas, we found out that any
non-greedy behavior significantly inhibits the agent's ability to
learn in the $2048$ game. Apparently, the inherently stochastic 2048
environment provides itself enough exploration.

\begin{algorithm}[h]
\centering{}\selectlanguage{american}%
\begin{center}
\begin{algorithmic}[1]
\Function{Learn}{}
    \While{\textbf{not} enough learning}
        \State $s_0 \gets \Call{InitialState}{ }$
        \State \Call{LearnFromEpisode}{$s_0$}
    \EndWhile
\EndFunction
\\
\Function{LearnFromEpisode}{$s_0$}
    \State $t = 0$
	\While {\textbf{not} \Call{Terminal}{$s_t$}}
		\State $a_t \gets \argmax_{a \in A(s_t)} \Call{Evaluate}{s_t, a_t}$
		\State $r_{t+1},s'_{t+1},s_{t+1} \gets \Call{MakeAction}{s_t, a_t}$
		\If{$t > 0$}
			\State \Call{TD($\lambda$)Update}{$r_{t+1} + V(s_{t+1}') - V(s'_t)$}
        \EndIf
        \State $t\gets t+1$
	\EndWhile
    \State \Call{TD($\lambda$)Update}{$-V(s'_t)$}
\EndFunction
\\
\Function{MakeAction}{$s, a$}
	\State $s', r \gets \Call{GetAfterstate}{s, a}$
	\State $s'' \gets \Call{AddRandomTile}{s'}$
	\Return $r,s',s''$
\EndFunction
\\
\Function{Evaluate}{$s, a$}
	\State $s', r \gets \Call{GetAfterstate}{s, a}$
	\Return $r + V(s')$
\EndFunction
\\
\State \textbf{parameters:} $\alpha,\lambda,h$
\Function{TD($\lambda$)Update}{$\delta_t$}
  \For{$k=t\textbf{ downto }\max(t-h,1)$}
    \For{$i=1\textbf{ to } m$}
\State $V_i[s'_k] \gets V_i[s'_k] + \frac{\alpha}{m}\delta_t\lambda^{t-k}$
    \EndFor
  \EndFor
\EndFunction

\end{algorithmic}
\par\end{center}\selectlanguage{english}%
\caption{\foreignlanguage{american}{\label{alg:AfterstateFramework}\foreignlanguage{english}{Framework
for TD($\lambda$) learning afterstate-value $n$-tuple networks.}}}
\end{algorithm}

\section{Algorithms and Experiments}

In this section, after presenting the common experimental setup, we
introduce, one by one, some existing and several novel algorithms
improving the afterstate learning framework introduced in Section
\ref{sec:TDL-NTuple}.

\subsection{Experimental Setup}

\subsubsection{Stopping Condition}

In $2048$, the length of an episode highly depends on the agent's
competence, which increases with the number of learning episodes.
This is why, in all the experiments in this paper, we limit the number
of actions an agent can make during its lifetime instead of the number
of episodes it can play. Unless stated otherwise, we give all the
methods the same computational budget of $10^{10}$ actions, which
corresponds to approximately $10^{7}$ learning episodes for the best
algorithms.

\subsubsection{Performance Measures}

We aimed at maximizing the expected score obtained by the agent. To
monitor the learning progress, every $2\times10^{8}$ actions, the
learning agent played $1000$ episodes using the greedy policy w.r.t.
value function. The average score we denote as \emph{$1$-ply performance}.
Since we were ultimately interested in combining the learned evaluation
function with a tree search, we also regularly checked how the agent
performs when coupled with expectimax \cite{michie1966game,russell1995artificial}
of depth $3$ (\emph{$3$-ply performance}). We played $300$ episodes
for this purpose.

\subsubsection{Computational Aspects}

All algorithms presented in this paper were implemented in Java. Experiments
were executed on a machine equipped with two $2.4$ GHz AMD Opteron\texttrademark{}
6234 CPUs and $64$ GB RAM. In order to take advantage of its $24$
cores, we applied the \emph{lock-free optimistic parallelism}, reducing
the learning time by a factor of $24$ comparing to the single-threaded
program.

Lock-free optimistic parallelism consists in letting all $24$ threads
play and learn $2048$ in parallel using shared data without any locks.
Since all threads simultaneously read from and write to the same $n$-tuple
network lookup tables, in theory, race conditions can occur. In practice,
however, since the $n$-tuple network can involve millions of parameters,
the probability of a race condition is negligible, and we have not
observed any adverse effects of this technique. 

Even with the parallelization, however, a single $2048$ learning
run takes, depending on the algorithm, $1$\textendash $7$ days to
complete. Thus, for statistics, we could repeat each algorithm run
only $5$ times.

\subsection{Automatic Adaptive Learning Rate}

One of the most critical parameters of temporal difference learning
is the learning rate $\alpha$, which is hard to setup correctly by
hand. This is why several algorithms that automatically set the learning
rate and adapt it over time have been proposed. For linear function
approximation, the list of online adaptive learning algorithms include
Beal's Temporal Coherence (\cite{beal1999temporal}), Dabney's $\alpha$-bounds
\cite{dabney2012adaptive}, Sutton's IDBD \cite{sutton1992adapting},
and Mahmood's Autostep \cite{mahmood2012tuning}. A thorough comparison
of these and other methods has recently been performed by Bagheri
et al. \cite{Bagheri2014Connect4} on Connect 4. It was shown that,
while the learning rate adaptation methods clearly outperform the
standard TD, the difference between them is not substantial, especially,
in the long run. That is why we evaluated only two of them here: the
simplest Temporal Coherence (TC($\lambda$)) \cite{beal1999temporal}
and the advanced, tuning-free Autostep , which was among the best
methods not only for Connect 4 but also for Dot and Boxes \cite{thill2015temporal}.
. Both of them, not only adjust the learning rate automatically but
maintain separate learning rates for each learning parameter.

\subsubsection{TC($\lambda$)}

\begin{algorithm}
\centering{}\selectlanguage{american}%
\begin{center}
\begin{algorithmic}[1]
\State \textbf{parameters:} $\beta,\lambda,h$
\Function{TC($\lambda$)Update}{$\delta_t$}
\For{$k=t\textbf{ downto }\max(1,t-h)$}
 \For{$i=1\textbf{ to } m$}
  \State $\alpha_i=\begin{cases} \frac{\left|E_i[s'_k]\right|}{A_i[s'_k]}, & \text{if } A_i[s'_k]\neq 0\\ 1, & \text{otherwise }\end{cases}$
  \State $V_i[s'_k] \gets V_i[s'_k] + \beta\frac{\alpha_i}{m}\delta_t\lambda^{t-k}$
  \State $E_i[s'_k]\gets E_i[s'_k] + \delta_t$
  \State $A_i[s'_k]\gets A_i[s'_k] + \left|\delta_t\right|$
 \EndFor
\EndFor
\EndFunction
\end{algorithmic}
\par\end{center}\selectlanguage{english}%
\caption{\label{alg:TC}TC($\lambda$) for the $n$-tuple network function
approximator.}
\end{algorithm}
Algorithm~\ref{alg:TC} shows the \textsc{TC($\lambda$)Update} function
which implements TC($\lambda$), replacing the \textsc{TD($\lambda$)Update}
in Alg.~\ref{alg:AfterstateFramework}. Except the afterstate-values
function $V$, the algorithm involves two other functions: $E$ and
$A$. For each afterstate $s'$, they accumulate the sum of errors
and the sum of absolute of errors, respectively (lines $7$-$8$).
Initially, the learning rate is $1$ for each learning parameter ($E[s']=A[s']=0$).
Then it is determined by the expression $\left|E_{i}[s']\right|/A_{i}[s']$
and will stay high when the two error accumulators have the same sign.
When the signs start to vary, the learning rate decreases. TC($\lambda$)
involves a meta-learning rate parameter $\beta$ (line 5). As with
$\alpha$ for TD($\lambda$), since we scale down the expression by
the number of $n$-tuples $m$, $\beta\in[0,1]$. 

For $E$ and $A$ we use $n$-tuple networks of the same architecture
as $V$.

\subsubsection{Autostep}

Autostep \cite{mahmood2012tuning} also uses separate learning rates
for each function approximator weight, but it updates it in a significantly
more sophisticated way than TC. Autostep extends Sutton's IDBD \cite{sutton1992adapting}
by normalizing the step-size update, setting an upper bound on it,
and reducing the step-size value when detected to be large. It has
three parameters: the meta-learning rate $\mu$, discounting factor
$\tau$, and initial step-size $\alpha_{init}$. Our implementation
of Autostep strictly followed the original paper \cite{mahmood2012tuning}.

\subsubsection{Results}

In the computational experiment, we evaluated three methods: TD($\lambda$),
TC($\lambda$) and Autostep, which differ by the number of parameters.
Due to the computational cost of the learning, we were not able to
systematically examine the whole parameter space and concentrated
on the learning rate $\alpha$ for TD($\lambda$), meta learning rate
$\beta$ for TD($\lambda$) and $\mu$ parameter for Autostep. Other
choices were made in result of informal preliminary experiments.

In these experiments, for TD($\lambda$) and TC($\lambda$), we found
that $\lambda=0.5$, $h=3$ work the best. Note that Autostep($\lambda$)
has not been defined by its authors, thus we used $\lambda=0$. We
also set $\alpha_{init}=1.0$ and $\tau=0.0001$, but found out that
the algorithm is robust to the two parameters, which complies with
its authors' claims \cite{mahmood2012tuning}.

Table~\ref{tab:LearningRateTuning} shows the results of the computational
experiment. The results indicate that, as expected TD($0.5$) is sensitive
to $\alpha$. Although TC($\lambda$) and Autostep automatically adapt
the learning rates, the results show that they are still significantly
susceptible to meta-learning parameters.

\begin{figure*}
\begin{centering}
\subfloat[Performance at $1$-ply]{\begin{centering}
\includegraphics[width=0.42\textwidth]{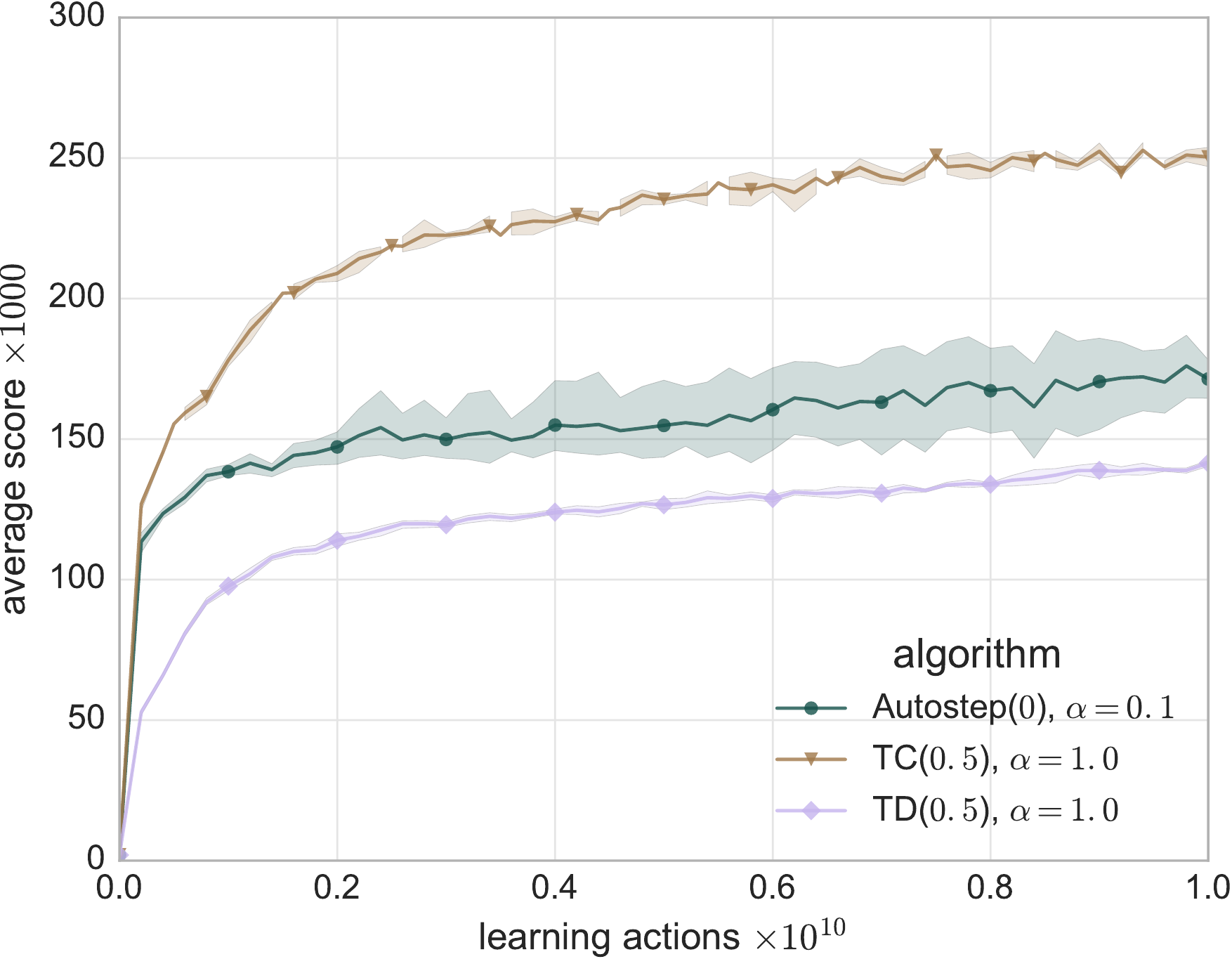}
\par\end{centering}
}\subfloat[Performance at $3$-ply]{\begin{centering}
\includegraphics[width=0.42\textwidth]{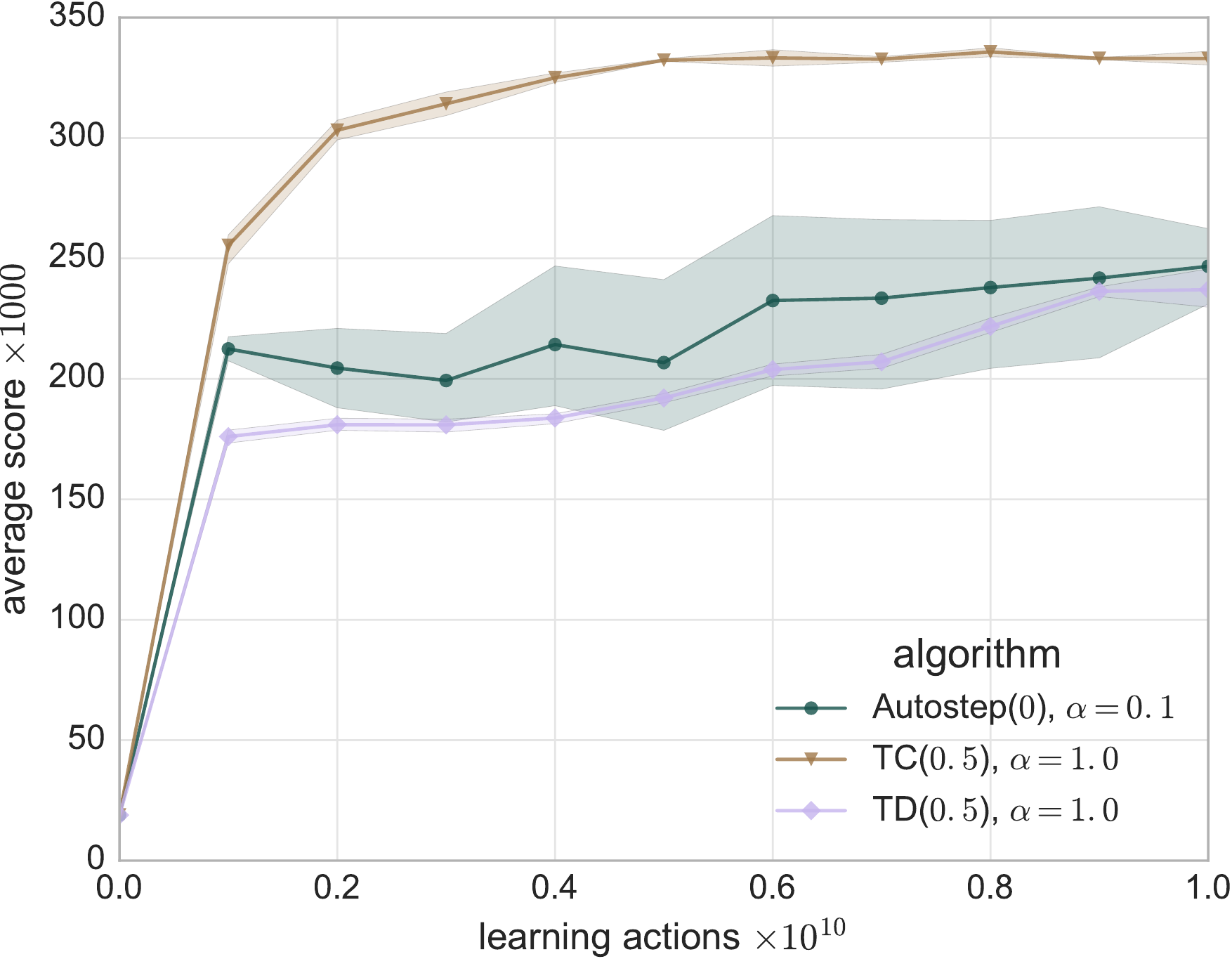}
\par\end{centering}
}
\par\end{centering}
\caption{\label{fig:TDTCAutostep}Comparison of the learning dynamics of TD($0.5$)
with $\alpha=1.0$, TC($0.5$) with $\beta=1.0$, and Autostep($0$)
with $\mu=0.1$. The transparent bands mean the $95\%$ confidence
interval.}
\end{figure*}

Figure~\ref{fig:TDTCAutostep} presents the comparison of the learning
dynamics for the best parameters found for the three algorithms considered.
Clearly, the automatic adaptation outperforms the baseline TD($0.5$).
Out of the two automatic learning rate adaptation methods, the simple
TC($0.5$) works better than the significantly more sophisticated
Autostep. Autostep starts as sharply as TC($0.5$), but then the learning
dynamics slows down considerably. Autostep is also characterized by
the higher variance (see the transparent bands). The poor performance
of Autostep might be due to the fact that it has been formulated only
for supervised (stationary) tasks and reinforcement learning for $2048$
makes the training data distribution highly non-stationary.

One disadvantage of both TC and Autostep is that since they involve
more $n$-tuple networks (e.g., for $E$ and $A$), they need more
memory access and thus increase the learning time. In our experiments
Autostep and TC($0.5$) took roughly $2$ and $1.5$ times more to
complete than TD($0.5$)\footnote{In this experiment we used the more efficient delayed-TD($\lambda$)
and delayed-TD($\lambda$) described in Section~\ref{subsec:Delayed-TD}.
The time differences would be much higher otherwise.}. However, the performance improvements shown in Fig.~\ref{fig:TDTCAutostep}
are worth this overhead. 

\begin{table}
\begin{centering}
\caption{\label{tab:LearningRateTuning}Influence of the (meta-)learning rate
parameters on the averages scores obtained by controllers trained
by TD($0.5$), TC($0.5$) and Autostep($0$) depending on the search
depth ($1$- or $3$-ply).}
\subfloat[TD($0.5$)]{\centering{}\def\arraystretch{1.2}
\sisetup{separate-uncertainty=true}
\begin{tabular}{
 S[table-format=1.3]
 S[table-format=6.0, table-figures-uncertainty=4, detect-weight]
 S[table-format=6.0, table-figures-uncertainty=5, detect-weight]
}
\toprule
$\alpha$ & {$1$-ply} & {$3$-ply} \\
\midrule
0.01 & 41085  \pm 265 & 99015   \pm 1318 \\
0.1  & 102130 \pm 1287 & 175301 \pm 1946  \\
0.5  & 130266 \pm 2136 & 196138 \pm 3864 \\
1.0  & \bfseries 141456 \pm 1050 & \bfseries 237026 \pm 8081 \\
\bottomrule
\end{tabular}}
\par\end{centering}
\begin{centering}
\subfloat[TC($0.5$)]{\centering{}\def\arraystretch{1.2}
\sisetup{separate-uncertainty=true}
\begin{tabular}{
 S[table-format=1.3]
 S[table-format=6.0, table-figures-uncertainty=4, detect-weight]
 S[table-format=6.0, table-figures-uncertainty=5, detect-weight]
}
\toprule
$\beta$ & {$1$-ply} & {$3$-ply} \\
\midrule
0.1  & 200588 \pm  1581 & 292558 \pm  4856  \\
0.5  & 244905 \pm  3811 & \bfseries 335076 \pm  1639 \\
1.0  & \bfseries 250393 \pm  3424 & 333010 \pm  2856 \\
\bottomrule
\end{tabular}}
\par\end{centering}
\centering{}\subfloat[Autostep with $\alpha_{init}=1.0$, $\tau=0.0001$]{\centering{}\def\arraystretch{1.2}
\sisetup{separate-uncertainty=true}
\begin{tabular}{
 S[table-format=1.3]
 S[table-format=6.0, table-figures-uncertainty=4, detect-weight]
 S[table-format=6.0, table-figures-uncertainty=5, detect-weight]
}
\toprule
$\mu$ & {$1$-ply} & {$3$-ply} \\
\midrule
0.001 & 90758 \pm  1789  & 146171 \pm  1023  \\
0.01  & 152737 \pm  6573 & \bfseries 249609 \pm 13142 \\
0.1   & \bfseries 171396 \pm  6662 & 246740 \pm 15469 \\
0.4   & 156112 \pm  8788 & 200391 \pm 18425 \\
0.7   & 33490 \pm   726  & 66557 \pm   659 \\
1.0   & 2546 \pm    62   & 5889 \pm   157 \\
\bottomrule
\end{tabular}}
\end{table}

\subsection{Delayed Temporal Difference\label{subsec:Delayed-TD}}

We have observed that the computational bottleneck of the learning
process lies in the access to the $n$-tuples lookup tables. This
is because the lookup tables are large and do not fit into the CPU
cache. In the TD($\lambda$) implementation presented in Alg.~\ref{alg:AfterstateFramework}),
the number of lookup table accesses is $O(mh)$, where $m$ is the
number of $n$-tuples and $h$ is the update horizon (lines $30$\textendash $33$).This
is because in each step $t$, the current prediction error $\delta_{t}$
is used to update the weights that were active during the $h+1$ last
states. Thus, each weight active in a given state is updated $h+1$
times until the state falls beyond the horizon. Notice, that the cumulative
error used as the update signal is a weighted sum of $h+1$ subsequent
prediction errors, i.e., $\Delta_{t}=\sum_{k=0}^{h}\delta_{t+k}\lambda^{k}$.
Our novel algorithm, delayed-TD($\lambda$), removes the dependency
on $h$.It operates as follows.

Delayed-TD($\lambda$) stores the last $h+1$ states $s_{t}$ and
errors $\delta_{t}$. Instead of instantly updating the weights active
in state $s_{t}$, it delays the update by $h$ steps. Then it updates
the weights by the (decayed) cumulative error~$\Delta_{t-h}$. This
way, the weights active in a given visited state are updated only
once. Since the vector of stored errors $\delta_{t}$ constitutes
a small, continuous memory block, it fits in the processor cache.
It makes its access time negligible compared to the time required
to access the main memory in which the large $n$-tuple network lookup
tables reside.

\begin{algorithm}
\centering{}\selectlanguage{american}%
\begin{center}
\begin{algorithmic}[1]
\State \textbf{parameters:} $\alpha,\lambda,h$
\Function{DelayedTD($\lambda$)Update}{$\delta_t$}
 \If{$t>h$}
  \State $\Delta_{t-h}=\sum_{k=0}^{h}{\delta_{t-h+k}\lambda^{k}}$
  \State \Call{ActualTDUpdate}{$s'_{t-h}, \Delta_{t-h}$}
 \EndIf
\EndFunction
\\
\Function{Finally}{}
  \For{$h'=h-1\textbf{ downto }0}$
    \State $\Delta_{t-h'}=\sum_{k=0}^{h'}{\delta_{t-h'+k}\lambda^{k}}$
    \State \Call{ActualTDUpdate}{$s'_{t-h'}, \Delta_{t-h}$}
  \EndFor
\EndFunction
\\
\Function{ActualTDUpdate}{$s', \Delta$}
  \For{$i=1\textbf{ to } m$}
    \State $V_i[s'] \gets V_i[s'] + \frac{\alpha}{m}\Delta$
  \EndFor
\EndFunction
\end{algorithmic}
\par\end{center}\selectlanguage{english}%
\caption{\foreignlanguage{american}{\label{alg:Delayed-TD}\foreignlanguage{english}{Delayed-TD($\lambda$)
for $n$-tuple network. The function \textsc{Finally} must be executed
as the last statement in the \textsc{LearnFromEpisode} function in
Alg.~\ref{alg:AfterstateFramework}.}}}
\end{algorithm}
\begin{algorithm}
\centering{}\selectlanguage{american}%
\begin{center}
\begin{algorithmic}[1]
\State \textbf{parameters:} $\beta,\lambda,h$
\Function{ActualTCUpdate}{$s',\Delta$}
 \For{$i=1\textbf{ to } m$}
  \State $\alpha_i=\begin{cases} \frac{\left|E_i[s']\right|}{A_i[s']}, & \text{if } A_i[s']\neq 0\\ 1, & \text{otherwise }\end{cases}$
  \State $V_i[s'] \gets V_i[s'] + \beta\frac{\alpha_i}{m}\Delta$
  \State $E_i[s']\gets E_i[s'] + \Delta$
  \State $A_i[s']\gets A_i[s'] + \left|\Delta\right|$
 \EndFor
\EndFunction
\end{algorithmic}
\par\end{center}\selectlanguage{english}%
\caption{\label{alg:DelayedTC}Delayed-TC($\lambda$) for the $n$-tuple network.
The functions \textsc{DelayedTC($\lambda$), Update($\delta_{t}$)
}and\textsc{ Finally }are analogous to the ones used in delayed-TD($\lambda$)
in Alg.~\ref{alg:Delayed-TD}.}
\end{algorithm}
The delayed versions of TD($\lambda$) and TC($\lambda$) are shown
in Alg.~\ref{alg:Delayed-TD} and \ref{alg:DelayedTC}. Notice that
since the actual update is delayed, it is done only when $t>h$. Also,
when the episode is finished, the algorithm needs to take care of
the most recently visited states. To this aim, the \textsc{Finally}
function is executed as the last statement in the \textsc{LearnFromEpisode}
function (Alg.~\ref{alg:AfterstateFramework}).

\begin{table}
\centering{}\caption{\label{tab:Delayed-TDvsTD}Delayed vs. standard versions of temporal
difference algorithms. We used $h=3$ and $\alpha=1.0$ for TD($0.5$)
and $\beta=1.0$ for TC($0.5$). The averages are accompanied by the
$95\%$ confidence interval.}
\begin{center}
\def\arraystretch{1.2}
\sisetup{separate-uncertainty=true, detect-weight}
\begin{tabular}{
 l
 S[table-format=5.0, table-figures-uncertainty=4]
 S[table-format=5.0, table-figures-uncertainty=4]
 S[table-format=1.2, table-figures-uncertainty=2]
}
\toprule
algorithm & {$1$-ply} & {$3$-ply} & {time [days]}\\
\midrule
TD(0.5)          & 140397 \pm  1974 & 234997 \pm  4717 & 0.26 \pm 0.01\\
delayed-TD(0.5)  & 141456 \pm  1050 & 237026 \pm  8081 & \bfseries 0.23 \pm 0.01\\
\midrule
TC(0.5)          & 248770 \pm  2296 & 333010 \pm  2856 & 1.16 \pm 0.03\\
delayed-TC(0.5)  & 250393 \pm  3424 & 335580 \pm  6299 & \bfseries 0.39 \pm 0.01\\
\bottomrule
\end{tabular}
\par\end{center}
\end{table}
The delayed temporal difference can be seen as a middle-ground between
the online and offline temporal difference rules \cite{Sutton1998Introduction-to}.
The computational experiment results shown in Table \ref{tab:Delayed-TDvsTD}
reveal no statistically significant differences between the standard
and delayed versions of temporal difference algorithms. On the other
hand, as expected, the delayed versions of the algorithms are significantly
quicker. Delayed-TC($0.5$) is roughly $3$ times more effective than
the standard TC($0.5$). The difference for TD($0.5$) is minor because
the weight update procedure is quick anyway and it is dominated by
the action selection (line $10$ in Alg.~\ref{alg:AfterstateFramework}).
In all the following experiments we will use the delayed version of
TC($0.5$).

\subsection{Multi-Stage Weight Promotion}

\subsubsection{Stages}

To further improve the learning process, we split the game into stages
and used a separate function approximator for each game stage. This
method has been already used in Buro's Logistello \cite{Buro1995Logistello:-A-S},
a master-level Othello playing program, and its variant with manually
selected stages has already been successfully applied to 2048 \cite{wu2014multi}.
The motivation for multiple stages for $2048$ results from the observation
that in order to precisely approximate the state-value function, certain
features should have different weights depending on the stage of the
game. The game has been split into $2^{g}$ stages, where $g$ is
a small positive integer. Assuming that the maximum tile to obtain
in $2048$ is $2^{15}=32\,768$, the ``length'' of each stage is
$l=2^{15+1-g}$. When $g=4$, $l=2^{12}=4096$. In this example, the
game starts in stage $0$ and switches to stage $1$ when $4096$-tile
is achieved, stage $2$ on $8192$-tile, stage $3$ when both $8192$
and $4096$ are on the board, stage $4$ on $16\,384$, and so on.

The value function for each stage is a separate $n$-tuple network.
Thus, introducing stages makes the model larger by a factor of $2^{g}$.
Thus, if $g=4$, the $33$-$42$ network requires $2^{4}\times4\times6^{16}=1\,073\,741\,824$
parameters.

We also need the multi-stage versions of $E$ and $A$ functions in
TC($\lambda$) algorithm, thus lines $5$-$7$ of Alg.~\ref{alg:TC}
become:

\begin{algorithmic}[1]
  \setcounter{ALG@line}{4}
  \State $V^{stage(s')}_i[s']\gets V^{stage(s')}_i[s'] + \beta\frac{\alpha_i}{n}\Delta$
  \State $E^{stage(s')}_i[s']\gets E^{stage(s')}_i[s'] + \Delta$
  \State $A^{stage(s')}_i[s']\gets A^{stage(s')}_i[s'] + \left|\Delta\right|$.
\end{algorithmic}

\subsubsection{Weight Promotion}

\begin{figure*}
\begin{centering}
\subfloat[$1$-ply performance]{\includegraphics[width=0.42\textwidth]{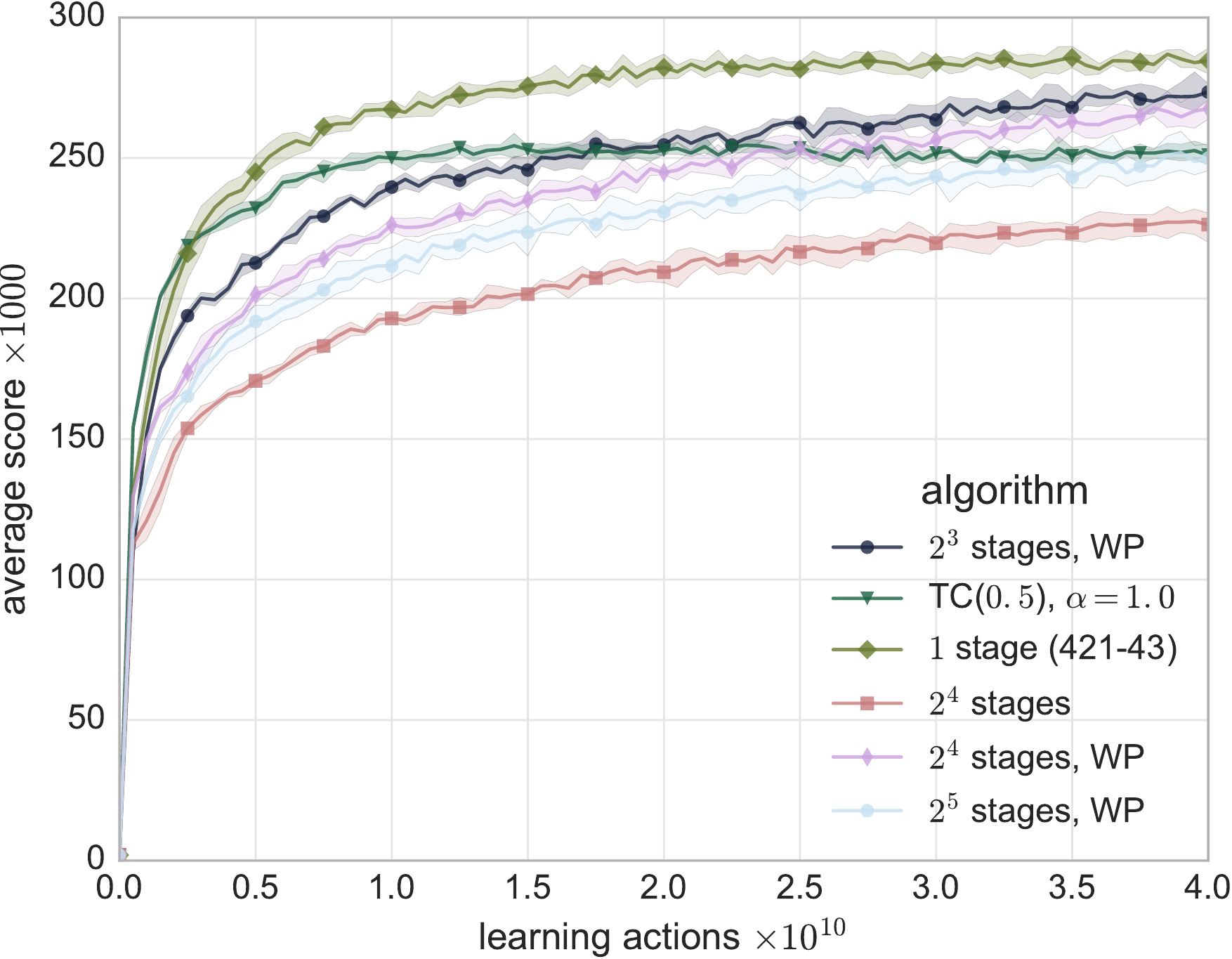}

}\subfloat[$3$-ply performance]{\includegraphics[width=0.42\textwidth]{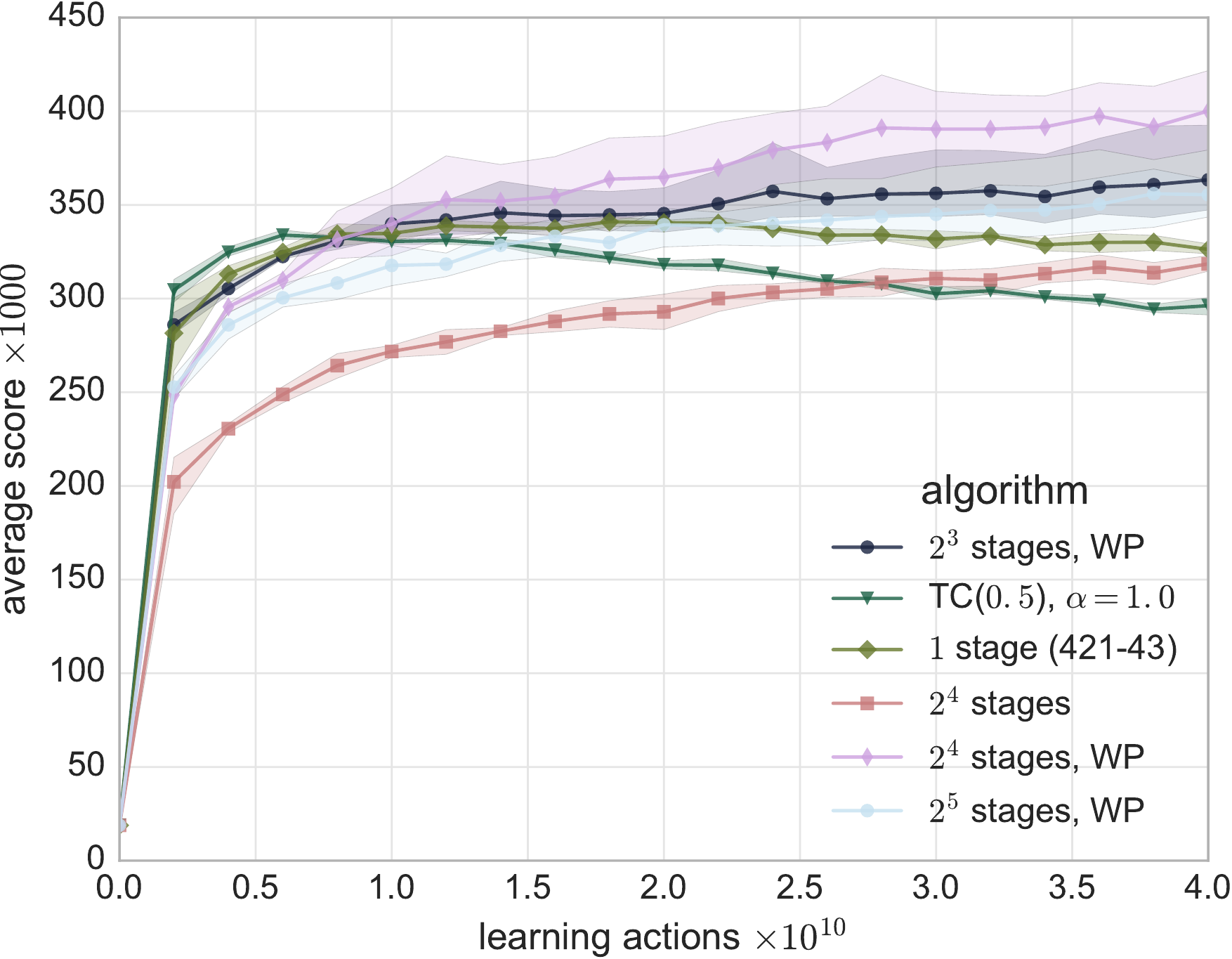}

}
\par\end{centering}
\centering{}\caption{\label{fig:MultiStateWeightProm}The effects of multiple stages and
weight promotion on learning performance.}
\end{figure*}
\begin{table}
\centering{}\caption{\label{tab:Multi-stage}Performance of algorithms involving multiple
stages and weight promotion (``WP'') after $4\times10^{10}$ actions.
All except the ``$1$ stage $421$-$43$'' variant used the $42$-$33$
network, trained with delayed-TC($0.5$), $\alpha=1.0$.}
\def\arraystretch{1.3}
\sisetup{separate-uncertainty=true}
\begin{tabular}{
 l
 S[table-format=5.0, table-figures-uncertainty=4, detect-weight]
 S[table-format=5.0, table-figures-uncertainty=3, detect-weight]
 S[table-format=1.2]
}
\toprule
{algorithm} & {$1$-ply} & {$3$-ply} & {time [days]}\\
\midrule
1 stage   &  251033 \pm  3483& 296207 \pm  4613& 1.62 \\
$2^4$ stages         &  226352 \pm  5433& 318534 \pm  4154& 2.22 \\
$2^3$ stages, WP     &  273510 \pm  3012& 363379 \pm 25276& 2.18 \\
$2^4$ stages, WP     &  267544 \pm  3707& \bfseries 400124 \pm 21739& 2.24 \\
$2^5$ stages, WP     &  250051 \pm  4790& 355413 \pm 10350& 2.42 \\
1 stage, $421$-$43$  &  \bfseries 284657 \pm  4605& 326267 \pm  2678& 1.76 \\
\bottomrule
\end{tabular}
\end{table}

The initial experiments with the multi-stage approximator revealed
that increasing the model by using the multi-stage approach actually
harms the learning performance. This is because, for each stage, the
$n$-tuple network must learn from scratch. As a result, the generalization
capabilities of the multi-stage function approximator are significantly
limited. In order to facilitate generalization without removing the
stages, we initialize each weight, upon its first access, to the weight
of the corresponding weight in the preceding stage (the weight from
a given stage is ``promoted'' to the subsequent one). To implement
this idea, the following lines are added in the\textsc{ Evaluate}
function between lines $25$ and $26$ of Alg.~\ref{alg:AfterstateFramework}:

\begin{center}
\selectlanguage{american}%
\begin{center}
\begin{algorithmic}[1]
 \setcounter{ALG@line}{25}
 \If{$V^{stage(s')}_i[s']$ not accessed \textbf{and} $stage(s')>0$}
    \State $V_i^{stage(s')}[s'] \gets V_i^{stage(s')-1}[s']$
 \EndIf
\end{algorithmic}
\par\end{center}\selectlanguage{english}%
\par\end{center}

In order to avoid adding additional data structures (memory and performance
hit), we implement the condition in line $21$ as checking whether
$V_{i}^{stage(s')}[s']$ equals $0$, heuristically assuming that
$V_{i}^{stage(s')}[s']=0$ until the first access.

\subsubsection{Results}

In order to evaluate the influence of the multi-stage approximator
and weight promotion, we performed a computational experiment with
six algorithm variants.

The results, shown in Table~\ref{tab:Multi-stage} and Fig.~\ref{fig:MultiStateWeightProm},
indicate that the multi-stage approximator involving $2^{4}$ stages
actually slows down the learning compared to the single stage baseline.
This is generally true for both $1$-ply and $3$-ply, but at $3$-ply,
the $2^{4}$-stages approximator, eventually, outperforms the $1$-stage
baseline. However, Fig.~\ref{fig:MultiStateWeightProm} shows that
it is due to the drop of performance of the $1$-stage approximator
rather than to the increase of the learning performance of the multi-stage
approach. After $1\times10^{10}$ actions, the $1$-stage approximator's
$3$-ply performance starts to decrease due to overfitting to $1$-ply
settings, which is used during the learning.

Essential to defeat the $1$-stage baseline is the weight promotion.
Although the algorithm variants with weight promotion still learn
at a slower pace than the baseline, they eventually outperform it
regardless of the search tree depth.

What is the optimal number of stages? The answer to this question
depends on the depth at which the agent is evaluated. The results
show that $2^{5}$ stages is overkill since it works worse than $2^{4}$
and $2^{3}$ stages. The choice between $2^{3}$ and $2^{4}$ is not
evident, though. For $1$-ply, the approximator involving $2^{3}$
stages learns faster and achieves a slightly better performance than
the one with $2^{4}$ stages. But for $3$-ply the situation is opposite
and $2^{4}$ clearly excels over $2^{3}$. Since we are interested
in the best score in absolute terms, we favor the $2^{4}$ stages
variant.

More stages involve more parameters to learn. As we already stated,
$2^{4}$ stages imply $16$ times more parameters than the $1$-stage
baseline. Is there a better way to spent the parameter budget? In
order to answer this question, we constructed a large single stage
$421$-$43$ network containing $5$ $7$-tuples of two shapes shown
in Fig.~\ref{fig:43-421}. The network is of a similar size to the
$2^{4}$ stages $42$-$33$ network, involving $5\times16^{7}=1\,342\,177\,280$
weights. Fig.~\ref{fig:MultiStateWeightProm} reveals that the new
network indeed excels at $1$-ply, but performs poorly at $3$-ply.
Interestingly, at $3$-ply, it not only does not improve over the
$42$-$33$ network but it also suffers from the $1$-ply overfitting
syndrome as it was the case with the $1$-stage baseline. 

Thus, it is tempting to conclude that multi-stage approximators are
immune to the $1$-ply overfitting syndrome. Notice, however, that
the overfitting happens only after the process converges and neither
of the multi-stage variants has converged yet.

Our best setup at $3$-ply involving $2^{4}$ stages and weight promotion
significantly improved over the $1$-stage baseline. Although, neither
the number of stages nor the weight selection did not negatively influence
the learning time, the new algorithm was found to require $4$ times
more actions to converge (and it still exhibits some growth potential).

\subsection{Carousel Shaping and Redundant Encoding}

As we have seen in the previous section, the $1$-ply performance
does not always positively correlate with the $3$-ply performance.
Some algorithms exhibit an overfitting to $1$-ply settings \textemdash{}
the better an agent gets at $1$-ply, the worse it scores at $3$-ply.

\subsubsection{Carousel Shaping}

Here we deal with $1$-ply overfitting by introducing \emph{carousel
shaping}, which concentrates the learning effort on later stages of
$2048$. This is motivated by the fact that only a small fraction
of episodes gets to the later stages of the game. Carousel shaping
tries to compensate this. Accurate estimates of the later stages of
the game are particularly important for $k$-ply agents, where $k>1$,
since they can look ahead further than the $1$-ply ones.

In carousel shaping, we use the notion of an\emph{ initial state of
a stage. }Any game's initial state is an initial state of stage $1$.
If two subsequent states $s'_{t-1}$, and $s'_{t}$ such that $stage(s'_{t-1})+1=stage(s'_{t})$,
$s'_{t}$ is an initial stage of $stage(s'_{t})$. A set of the last
$1000$ visited initial states of each stage is maintained during
the learning.

The carousel shaping algorithm starts from an initial state of stage
$1$ After each learning episode, it proceeds to the subsequent stage.
It starts each episode with a randomly selected initial state of the
current stage. It restarts from stage $1$, when it hits an empty
set of initial states (see Alg.~\ref{alg:CarouselShaping}). This
way, the learning more often visits the later stages of the game.

Note that this technique is different from the one introduced for
$2048$ by Wu et al. \cite{wu2014multi}. Their algorithm learns the
\emph{n}-tuple networks stage by stage. After some number of learning
episodes for a given stage, it moves to the next one and never updates
the weights for the previous one again. The state-value function learned
this way may not be accurate. New information learned in a given stage
does not backpropagate to the previous stages. This might lead to
suboptimal behavior.

\begin{algorithm}
\centering{}\selectlanguage{american}%
\begin{center}
\begin{algorithmic}[1]
\Function{CarouselShaping}{}
 \State $stage \gets 1$
 \State $initstates[x] = \emptyset$ \textbf{for} $x\in\{1\dots 2^g\}$
 \While{\textbf{not} enough learning}
   \If{$stage = 1$}
     \State $s \gets \Call{InitialState}{ }$
   \Else
     \State $s \gets \Call{RandomChoice}{initstates[stage]}$
   \EndIf
   \State \Call{LearnFromEpisode}{$s$} \Comment{Updates $initstates$}
   \State $stage \gets stage + 1$
   \If{$stage > 2^g$ \textbf{or} $initstates[stage]=\emptyset$}
     \State $stage \gets 1$
   \EndIf
 \EndWhile
\EndFunction
\end{algorithmic}
\par\end{center}\selectlanguage{english}%
\caption{\foreignlanguage{american}{\label{alg:CarouselShaping}\foreignlanguage{english}{Carousel shaping.}}}
\end{algorithm}

\subsubsection{Redundant Encoding}

The last technique we apply to $2048$ is \emph{redundant encoding}.
We define redundant encoding as additional features to the function
approximator that does not increase its expressiveness. Although the
new features are redundant from the point of view of the function
approximator, they can facilitate learning by quicker generalization.
For the $n$-tuple network, redundant encoding consists in adding
smaller $n$-tuples, which are comprised in the larger ones. In our
experiments, we first extended our standard $42$-$33$ network by
$4$-tuples of two shapes: straight lines of length $4$ and $2\times2$
squares denoted as ``$22$''. Finally, we added straight lines of
length $3$ (denoted as ``$3$'') (see Fig.~\ref{fig:NTN-architectures}).

\subsubsection{Results}

\begin{table*}
\centering{}\caption{\label{tab:CSRE}Performance of algorithms involving carousel shaping
(``CS'') and redundant encoding after $4\times10^{10}$ actions.
All variants used the $42$-$33$ network with $2^{4}$ stages and
weight promotion, trained with delayed-TC($0.5$), $\alpha=1.0$.}
\def\arraystretch{1.2}
\sisetup{
  separate-uncertainty=true,
}
\begin{tabular}{
 l
 S[table-format=6.0, table-figures-uncertainty=5, detect-weight]
 S[table-format=6.0, table-figures-uncertainty=5, detect-weight]
 S[table-format=2.0]
 S[table-format=1.2, detect-weight]
}
\toprule
{algorithm} & {$1$-ply} & {$3$-ply} & {\# $n$-tuples} & {time [days]}\\
\midrule
42-33 (baseline) & 265435\pm8511  & 393289\pm26496 & 5  & 2.35 \\
42-33, CS        & 258616\pm5784  & 432701\pm13005 & 5  & 2.78 \\
42-33-4-22, CS   & 256569\pm19975 & 457754\pm6256 & 10 & 4.63 \\
42-33-4-22-3     & \bfseries 319433\pm2708  & 469779\pm10920 & 12 & 5.06 \\
42-33-4-22-3, CS & 311426\pm9633  & \bfseries 491398\pm19458 & 12 & 5.47 \\
\bottomrule
\end{tabular}
\end{table*}
The results of applying carousel shaping and redundant encoding are
shown in Table.~\ref{tab:CSRE}. Clearly, the effect of carousel
shaping on $1$-ply performance is consistently slightly negative.
This is less important, however, since we can see that the method
significantly improves the $3$-ply performance regardless of the
network architecture.

We can also see that the best algorithm involves the redundant $4$-tuples
and $3$-tuples. Compared to the standard $42$-$33$ network, the
redundant $4$-tuples do not help at $1$-ply, but they make a difference
at $3$-ply. The redundant $3$-tuples show a significant improvement
both at $1$- and $3$-ply. The figure also indicates that there is
a synergetic effect of redundant encoding and carousel shaping, especially
at $3$-ply, where the algorithm using both techniques scores nearly
$500\,000$ points on average.

A downside of redundant encoding is that it makes the learning slower
due to involving more $n$-tuples, which results in more table lookups
for each state evaluation. Table~\ref{tab:CSRE} shows that the learning
time is roughly proportional to the number of $n$-tuples involved.
Notice, however, that the redundant $n$-tuples are required only
during the learning. After the learning is finished, the lookup tables
corresponding to the redundant $n$-tuples can be integrated into
the lookup tables of the n-tuples that contain the redundant ones.
This way, the resulting agent has the same performance as if it was
trained without redundant encoding.

\section{The Best $n$-Tuple Network}

\subsection{Expectimax Tree Search}

The best $n$-tuple network in the experiments has been obtained using
TC($0.5$) with $\alpha=1.0$ with redundant encoding (the $42$-$33$-$22$-$4$-$3$
network), $2^{4}$ stages, with weight promotion, and carousel shaping.
Here we evaluate it thoroughly.

Table \ref{fig:PerfBest} shows how the performance of the network
depends on how deep expectimax it is allowed to search the game tree.
For the experiment, the player was allowed to use either constant
depth ($1$-ply, $2$-ply, $3$-ply, or $5$-ply) or constant time
per move (iterative deepening with $1$ ms, $50$ ms, $100$ ms, $200$
ms or $1000$ ms). In the latter case, the agent could go deeper than
$5$-ply if the time allows, which happen especially in the end-games,
in which the branching factor gets smaller. Transposition tables were
used to improve the performance of expectimax. The program was executed
on Intel i7-4770K CPU @ $3.5$ GHz, single-threaded.

Table \ref{fig:PerfBest} shows that, generally, the deeper the search
tree, the better its average score is. It is also clear, however,
that the CPU time can be spent more effectively if we limit the time
per move rather than the search tree depth (compare the results for
$5$-ply vs. $100$ ms). We can also see that increasing the time
per move allows to make better decisions, which renders in better
scores.

\subsection{Comparison with Other $2048$ Controllers}

Table~\ref{fig:OtherPlayers} shows a performance comparison of our
best controller with other published $2048$ players. Clearly, our
player outperforms the rest. It is important to show how large the
gap between our controller and the runner-up is (which, apart from
hand-made features, also uses the $42$-$33$ network). Our $n$-tuple
network outranks it regardless of the expectimax settings: at $100$
ms per move and $3$-ply depth (the Yeh et. al's controller also uses
$3$-ply). Moreover, our network is even slightly better than the
runner-up at $2$-ply, when it can play more than $40$ times faster.
This confirms the effectiveness of the learning methodology developed
in this paper.

\begin{table*}
\centering{}\caption{\label{fig:PerfBest}Performance of the best $n$-tuple network with
respect to the search limit (constant depth or time limit per move).}
\begin{center}
\def\arraystretch{1.2}
\sisetup{separate-uncertainty=true, detect-weight}
\begin{tabular}{
 r
 S[table-format=6.0, table-figures-uncertainty=5]
 S[table-format=2.0]
 S[table-format=2.0]
 S[table-format=3.0]
 S[table-format=4.0]
 S[table-format=6.0]
}
\toprule
Search limit & {Average score} & {$32\,768$ [\%]} & {$16\,384$ [\%]} & {$8192$ [\%]} & {\# games} & {Moves/s} \\
\midrule
1-ply  & 324710 \pm 11043 & 19 & 68 &  90 & 1000 & 258371 \\
2-ply  & 457079 \pm 11112 & 34 & 91 &  99 & 1000 & 20524 \\
3-ply  & 511759 \pm 12021 & 50 & 92 &  99 & 1000 & 1484 \\
5-ply  & 545833 \pm 21500 & 54 & 97 & 100 & 300  & 16 \\
\midrule
1 ms   & 527099 \pm 11486 & 54 & 95 & 100 & 1000 & 916 \\
50 ms  & 576655 \pm 20839 & 62 & 97 & 99  & 300  & 20 \\
100 ms & 589355 \pm 20432 & 65 & 96 & 100 & 300  & 10 \\
200 ms & 591380 \pm 21870 & 67 & 97 & 99  & 300  & 5 \\
1000 ms & \bfseries 609104 \pm 38433 & 70 & 97 & 98 & 100 & 1\\
\bottomrule
\end{tabular}
\par\end{center}
\end{table*}

\begin{table*}
\centering{}\caption{\label{fig:OtherPlayers}Comparison with other methods.}
\begin{center}
\def\arraystretch{1.2}
\sisetup{separate-uncertainty=true}
\begin{tabular}{
 l
 S[table-format=6.0]
 S[table-format=2.0]
 S[table-format=6.0]
 l
}
\toprule
{Authors} & {Average score} & {$32\,768$ [\%]} & {Moves/s} & Method\\
\midrule
Szubert \& Ja\'skowski \cite{Szubert2014_2048} & 99916& 0& 330000& $n$-tuple network, TD($0$), $1$-ply\\
Oka \& Matsuzaki \cite{kazuko2016_2048} & 234136& 3& 88000& $n$-tuple network, TD($0$), $1$-ply\\
Wu et al. \cite{wu2014multi} & 328946 & 11 & 300 & multi-stage TD($0$), $3$-ply\\
Xiao et al. \cite{nneonneo} & 442419 & 32 & 3 & hand-made features, CMA-ES, adaptive depth\\
Yeh et al. \cite{Yeh2016_2048} & 443526 & 32 & 500& $n$-tuple network, hand-made features, $3$-ply\\
\midrule
\multirow{5}{*}{This work} & 324710 & 19 & 258371 & $n$-tuple network, $1$-ply \\
& 457079 & 34 & 20524 & $n$-tuple network, $2$-ply \\
& 511759 & 50 & 1464 & $n$-tuple network, $3$-ply \\
& 527099 & 54 & 916 & $n$-tuple network, $1$ ms/move\\
& 609104 & 70 & 1 & $n$-tuple network, $1000$ ms/move\\
\bottomrule
\end{tabular}
\par\end{center}
\end{table*}

\section{Discussion and Conclusions}

In this paper, we presented how to obtain a strong $2048$ controller.
It scores more than $600\,000$ points on average and achieves the
$32\,768$-tile in $70\%$ of games, which is significantly more than
any other $2048$ computer program to date. It is also unlikely that
any human player can achieve such level of play. At least because
some episodes could require as much as $30\,000$ moves to finish,
which makes human mistakes inevitable.

In order to obtain this result, we took advantage of several existing
and introduced some new reinforcement learning techniques (see Table
\ref{tab:Summary}). First, we applied temporal coherence (TC) learning
that turned out to significantly improve the strength of the player
compared to the vanilla temporal difference (TD) learning. Second,
we used a different function approximator for each stage of the game
but have shown that it pays off only when additional generalization
methods are applied. For this aim, we introduced the weight promotion
technique. Third, we demonstrated that the controller can be further
improved when carousel shaping is employed. It makes the ``later''
stages of the game visited more often. It worsens the controller at
$1$-ply, but improves it at $3$-ply, which we really care about.
Finally, it was the redundant encoding that boosted the learning effectiveness
substantially. Redundant encoding facilitates generalization and,
as a result, the learning speed and the final controller score. Although
it slows the learning, the redundant $n$-tuples can be incorporated
into the non-redundant ones when the learning is finished, making
the final agent quicker.

\begin{table*}
\centering{}\caption{\label{tab:Summary}Summary of the performance improvements obtained
by introducing subsequent techniques}
\def\arraystretch{1.2}
\sisetup{
  separate-uncertainty=true,
}
\begin{tabular}{
 l
 S[table-format=6.0, table-figures-uncertainty=5, detect-weight]
 S[table-format=6.0, table-figures-uncertainty=5, detect-weight]
 S[table-format=2.0]
 S[table-format=1.2, detect-weight]
}
\toprule
{algorithm} & {$1$-ply} & {$3$-ply} & {learning days} \\
\midrule
TD($0.5$) & 141456 \pm 1050 & 237026 \pm 8081 & 0.26 \pm 0.01 \\
TC($0.5$) & 248770 \pm  2296 & 333010 \pm  2856 & 1.16 \pm 0.03 \\
delayed-TC($0.5$) & 250393 \pm  3424 & 335580 \pm  6299 & 0.39 \pm 0.01 \\
$+$ $2^4$ stages \& Weight Promotion &  267544 \pm  3707 & 400124 \pm 21739 & 2.24 \\
$+$ Redundant Encoding & 319433\pm2708  & 469779\pm10920 & 5.06 \\
$+$ Carousel Shaping & 311426 \pm 9633  & \bfseries 491398\pm19458 & 5.47 \\
\bottomrule
\end{tabular}
\end{table*}

We also showed some techniques to lessen the computational burden.
We introduced delayed versions of the temporal difference update methods
and demonstrated that it can reduce the learning time of temporal
coherence (TC) several times making the state-value function update
mechanism vastly independent of the decay parameter $\lambda$. Secondly,
we proved the practical efficacy of the lock-free optimistic parallelism,
which made it possible to utilize all CPU cores during the learning.

From the broader point of view, although we concentrated on a specific
problem of the game $2048$, the techniques we introduced are general
and can be applied to other discrete Markov decision problems, in
which $n$-tuple networks or, more broadly, tile encodings, can be
effectively used for function approximation.

In this context, it is worth to emphasize the role of the (systematic)
$n$-tuple networks for the overall result. Although they do not scale
up to state spaces with large dimensionality \cite{Jaskowski2015sztetris},
they have undeniable advantages for state spaces of moderate sizes
and dimensions. They provide nonlinear transformations and computational
efficiency. As we demonstrated, $n$-tuple networks can have billions
of parameters, but at the same time, only a tiny fraction of them
is involved to evaluate a given state or to update the state-value
function. On the one hand, the enormous number of parameters of $n$-tuple
network can be seen as a disadvantage as it requires a vast amount
of computer memory. On the other hand, this is also the reason that
the lock-free optimistic parallelism can work seamlessly.

From the $2048$ perspective, this work can be extended in several
directions. First of all, although $n$-tuple networks allowed us
to obtain a very competent $2048$ player, we believe they do not
generalize too well for this problem. They rather learn by heart.
Humans generalize significantly better. For example, for two board
states, one of which has tiles with doubled values, a human player
would certainly use similar, if not the same, strategy. Our $n$-tuple
network cannot do that. This is why the learning takes days. So the
open question is, how to design an efficient state-value function
approximator that generalizes better? Actually, it is not clear whether
this is possible at all since for the above example a state-value
approximator is supposed to return two different values. We speculate
that better generalization (and thus, faster learning) for $2048$
may require turning away from state-value functions and learning policies
directly. A directly learned policy does not need to return numbers
that have absolute meaning, thus, in principle, can generalize over
to the boards mentioned in the example.

The second direction to improve the $2048$ controller performance
involves including some hand-designed features. Although local features
would probably not help since the $n$-tuple networks involve all
possible local features, what our function approximator lacks are
global features like the number of empty tiles on the board. 

Some of the introduced techniques could also be improved or fine-tuned
at least. For example, we did not experiment with carousel shaping.
It is possible to parametrize it in order to control how much it puts
the learning attention to the ``later'' stages. The question about
how to effectively involve exploration is also open. No exploration
mechanism we tried worked, but the lack of exploration may, eventually,
make the learning prone to stuck in local minima, so an efficient
exploration for $2048$ is another open question. Moreover, the learning
rate adaptation mechanism of temporal coherence is somewhat trivial
and arbitrary. Since the more sophisticated Autostep failed, there
can be an opportunity for a new learning rate adaptation method.

Last but not least, we demonstrated how the learned evaluation function
can be harnessed by a tree search algorithm. We used the standard
expectimax with iterative deepening with transposition tables, but
some advanced pruning \cite{schadd2009chanceprobcut} could potentially
improve its performance.\footnote{The Java source code used to perform the learning experiments shown
in this paper is available at \url{https://github.com/wjaskowski/mastering-2048}.
A more efficient C++ code for running the best player along with the
best $n$-tuple network can be found at \url{https://github.com/aszczepanski/2048}.}

\subsection*{Acknowledgments}

The author thank Marcin Szubert for his comments to the manuscript
and Adam Szczepa\'{n}ski for implementing an efficient C++ version
of the $2048$ agent. This work has been supported by the Polish National
Science Centre grant no. DEC-2013/09/D/ST6/03932. The computations
were performed in Poznan Supercomputing and Networking Center.

\bibliographystyle{plain}
\bibliography{wjaskowski,library,rl,all}

\end{document}